\documentclass{article}
\usepackage{spconf}
\usepackage{signalpreamble}

\usepackage{tikz}
\usepackage{pgfplots}
\pgfplotsset{compat=newest}

\newacronym[plural=GDSs]{GDS}{GDS}{graph distribution-valued signal}
\newacronym{GSP}{GSP}{Graph signal processing}
\newacronym{GFT}{GFT}{graph Fourier transform}
\newacronym{GSO}{GSO}{graph shift operator}
\newacronym{GDS-FT}{GDS-FT}{GDS Fourier transform}
\newacronym{CDF}{CDF}{cumulative distribution function}

\title{Graph Distribution-valued Signals: A Wasserstein Space Perspective}
\name{Yanan Zhao$^{1}$, Feng Ji$^{1}$, Xingchao Jian$^{2}$, Wee Peng Tay$^{1}$}
\address{%
\fontsize{10}{10}\selectfont
$^{1}$School of Electrical and Electronic Engineering, Nanyang Technological University, Singapore\\
\fontsize{10}{10}\selectfont
$^{2}$School of Computation, Information and Technology, Technical University of Munich, Germany\\
\fontsize{10}{10}\selectfont
Emails: \{YANAN002, xingchao001\}@e.ntu.edu.sg, \{jifeng, wptay\}@ntu.edu.sg}

\begin{document}
\ninept

\maketitle

\begin{abstract}
We introduce a novel framework for graph signal processing (GSP) that models signals as graph distribution-valued signals (GDSs), which are probability distributions in the Wasserstein space. This approach overcomes key limitations of classical vector-based GSP, including the assumption of synchronous observations over vertices, the inability to capture uncertainty, and the requirement for strict correspondence in graph filtering. By representing signals as distributions, GDSs naturally encode uncertainty and stochasticity, while strictly generalizing traditional graph signals. We establish a systematic dictionary mapping core GSP concepts to their GDS counterparts, demonstrating that classical definitions are recovered as special cases. The effectiveness of the framework is validated through graph filter learning for prediction tasks, supported by experimental results.
\end{abstract}


\section{Introduction}

Graphs provide a natural framework for modeling complex systems in domains such as social networks, transportation infrastructures, and sensor arrays. In classical \gls{GSP} \cite{Shu13, San14, Ort18, ortega_2022, Girault2018, Ji19}, a graph signal is represented as a vector, with each entry corresponding to a node’s value. This representation enables the use of linear operators, including the \gls{GFT}, convolution, and graph filters \cite{Shu13, Ort18, ortega_2022}, to analyze signal structure and inter-node relationships.

Despite its widespread adoption, the vector-based \gls{GSP} framework faces two key limitations: 
(i) \emph{Complete observation assumption.} Classical GSP typically presumes that signals from all nodes are observed simultaneously, enabling the application of operators such as the \gls{GFT} to a complete signal vector. In practice, however, data collection is often asynchronous or incomplete across nodes \cite{Hua20, Ji19, JiTayOrt23}, making this assumption unrealistic.
(ii) \emph{Strict correspondence requirement.} While statistical \gls{GSP} methods \cite{PerVan:J17,MarSegLeuRib:J17,JiaTay:J22,JiaTayEld24,KroRouEld:J22,SagRou23} introduce randomness via random variables, graph filters are still defined as vector-to-vector mappings and require paired source and target signals, e.g., $\{(\bx_i, \by_i)\}_{i=1}^m$, with the filter $\bF$ learned so that $\bF(\bx_i) \approx \by_i$. In prediction tasks, this enforces a rigid one-to-one correspondence between predictor and response signals. In reality, alignment is often imperfect due to shifts, overlaps, periodic patterns, or missing data, limiting the practical applicability of such approaches.

To address these challenges, we propose modeling graph signals as probability distributions rather than as random vectors or sample realizations. Unlike statistical \gls{GSP}, which typically operates on samples and prior knowledge of the underlying distribution \cite{KroRouEld:J22,SagRou23,JiaGolJi25} or statistical properties such as stationarity \cite{PerVan:J17,MarSegLeuRib:J17,JiaTay:J22}, our framework treats the entire probability distribution as the primary signal object. When the distribution model is known, our approach can be viewed as a \emph{zero-shot} filter learning paradigm—directly optimizing filters in the distribution space without requiring sample data. If the distribution is unknown, samples are first used to estimate the underlying distribution, which then serves as the basis for filter learning. This contrasts with statistical \gls{GSP}, where filters are learned directly from samples. Note that most statistical \gls{GSP} methods suffer from the two limitations mentioned above, as they still rely on vector-based signal representations. In our framework, we replace the classical vector space of graph signals with the Wasserstein space \cite{Vil09,Kol56,Dur19}, whose elements are probability measures on $\mathbb{R}^{N}$. We refer to such a signal as a \emph{\gls{GDS}}.

\begin{figure}[!t]
    \centering
    \begin{subfigure}[b]{0.48\columnwidth}
        \centering
        \includegraphics[width=1.12\textwidth, trim={3cm 9cm 1.5cm 8cm}, clip]{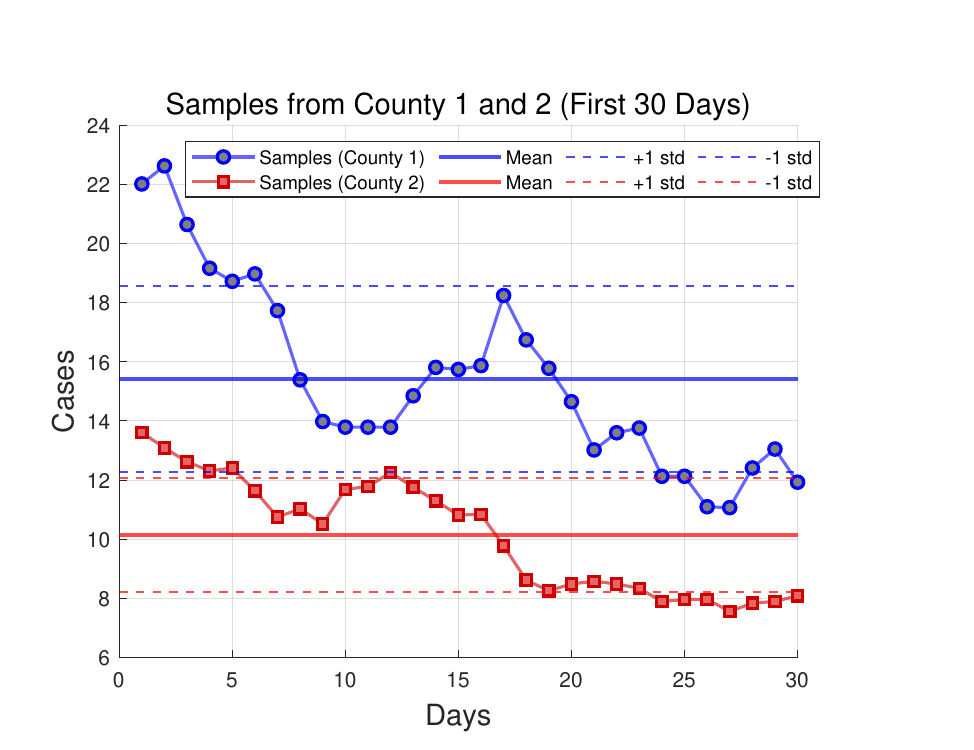}
        \vspace{-6mm}
        \caption{Daily cases from two counties}
        \label{fig.covid_samples}
    \end{subfigure}
    \vspace{-4mm}
    \begin{subfigure}[b]{0.48\columnwidth}
        \centering
        \includegraphics[width=1.12\textwidth, trim={3cm 9cm 1.5cm 8cm}, clip]{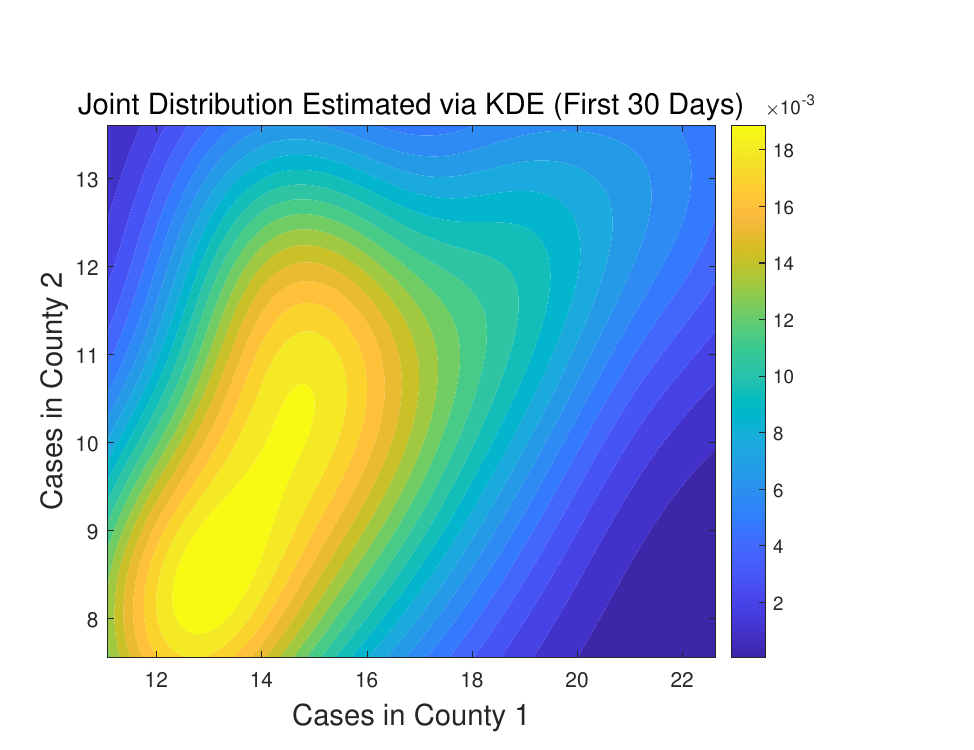}
        \vspace{-6mm}
        \caption{KDE-based joint distribution}
        \label{fig.covid_distribution}
    \end{subfigure}
    \caption{(a) COVID-19 daily cases from two counties over 30 days with mean and $\pm1$ standard deviation. (b) Joint distribution estimated via KDE, showing the trajectories in (a) as realizations from an underlying distribution in the GDS framework.}
    \label{fig:samples_vs_distributions}
    \vspace{-3mm}
\end{figure}

To illustrate the distinction, consider a COVID-19 dataset reporting daily cases across 58 counties from 2020 to 2022. In classical \gls{GSP}, each day's graph signal is a vector containing the reported cases for all counties, which is a single snapshot. In contrast, the \gls{GDS} framework aggregates these daily vectors as samples from an underlying joint distribution, estimates this distribution, and treats it as the graph signal. This approach encodes both variability and inter-county dependencies. \Cref{fig:samples_vs_distributions} highlights this difference: \cref{fig.covid_samples} presents daily case counts for two counties over 30 days (vector-valued samples), while \cref{fig.covid_distribution} shows the estimated joint distribution via kernel density estimation (KDE), reflecting the distributional perspective. Modeling signals as distributions naturally captures stochasticity (inherent variability) and uncertainty (arising from incomplete or noisy observations). Importantly, the Wasserstein space strictly generalizes the classical vector space: traditional graph signals correspond to single-point mass distributions and are thus special cases of \glspl{GDS}. This extension enables principled handling of uncertainty and richer modeling of graph-structured data.

A drawback of the \gls{GDS} framework is that additional data may be required to accurately estimate distributions, especially in high-dimensional settings. In practice, applications often do not require full distributional knowledge. Instead, one can assume a parametric model like a Gaussian copula to capture key statistical properties. In some other applications like node classification, deep learning models can be employed to learn a proxy distribution given by the softmax output of the model for the node labels, with the GDS framework applied to filter these node distributions for downstream tasks \cite{JiLeeMen23,JiZhaLee25}. Another drawback is increased computational complexity due to operations in the Wasserstein space. However, this can be mitigated by working with tractable families of distributions, as demonstrated in our numerical experiments.

The central objective of this work is to construct a conceptual dictionary that systematically maps the key operations of \gls{GFT} and convolution from classical GSP to their analogues in the proposed \gls{GDS} framework. This mapping enables the direct translation of vector-based models and algorithms into the distribution-valued setting, facilitating principled extensions of existing GSP theory. The main contributions are summarized as follows:
\begin{itemize}
    \item We introduce the \gls{GDS} framework, modeling graph signals as probability distributions in the Wasserstein space to naturally encode uncertainty and stochasticity.
    \item We develop a systematic correspondence between foundational GSP concepts and their GDS counterparts, demonstrating that classical definitions are recovered as special cases.
    \item We propose a graph filter learning approach within the GDS framework and validate its effectiveness through experimental results.
\end{itemize}

\emph{Notations.} Scalars and scalar-valued functions are denoted by plain lowercase letters (e.g., $x$), while vectors and vector-valued functions are represented by bold lowercase letters (e.g., $\bx$). Matrices are denoted by bold uppercase letters (e.g., $\bA$). The set of real number is denoted by $\Real$. The transpose operation is denoted by $\parens{}\T$, and $\Tr(\cdot)$ denotes the trace operator.

\section{Wasserstein spaces and \glspl{GDS}}

Throughout this paper, we consider a simple undirected graph $\calG=(\calV,\bA)$ with $|\calV|=N$ vertices and weighted edges represented by the adjacency matrix $\bA=[a_{ij}]_{i,j\in\calV}$. An entry $a_{ij}>0$ indicates that vertices $i$ and $j$ are connected, while $a_{ij}=0$ otherwise. We assume that $\calG$ is connected, weighted, and has no self-loops. The corresponding graph Laplacian is defined as $\bL_{\calG}=\bD-\bA$, where $\bD$ is the degree matrix. We denote by $\bS_{\calG}$ a generic graph shift operator, such as adjacency matrix $\bA$ or the Laplacian $\bL_{\calG}$. The \gls{GFT} of a graph signal $\bx\in\bbR^N$ is given by $\widehat{\bx}=\bU\T_{\calG}\bx$, where $\bU_{\calG}$ is the unitary matrix whose columns are the eigenvectors of the chosen \gls{GSO}. 

A traditional graph signal is a vector $\bx \in \bbR^N$ that assigns a real value to each vertex. From a probabilistic perspective, this can be viewed as the Dirac measure $\delta_{\bx}$. This motivates a natural generalization in which graph signals are modeled as probability measures on $\bbR^N$, residing in the Wasserstein space \cite{Vil09}. This probabilistic formulation unifies deterministic signals and their statistical descriptions, providing a foundation for the proposed \gls{GDS} framework.

\begin{Definition}[Wasserstein space]
\label{def.wasse.space}
 Let $(\calX, d)$ be a metric space and $\calP(\calX)$ the set of Borel probability measures on $\calX$, i.e., $\calP(\calX):=\set{\mu \given \mu~\text{is a probability measure on}~(\calX, \calB(\calX))}$, where $\calB(\calX)$ is the Borel $\sigma$-algebra generated by open sets of $\calX$. For $p \ge 1$, the Wasserstein space of order $p$ is
 \begin{align*}
     \calP_{p}(\calX) = \set*{\mu \in \calP(\calX)\given \int_{\calX} d(x_{0},x)^{p}\ud\mu(x) < \infty}
 \end{align*}
 for some $x_{0}\in \calX$. The space is equipped with the $p$-Wassertain distance: for $\mu_1,\mu_2 \in \calP_{p}(\calX)$,
 \begin{align*}
     W_{p}(\mu_{1},\mu_{2}):= \left(\inf_{\gamma \in \Gamma(\mu_{1},\mu_{2})}\int_{\calX \times \calX}d(x,y)^{p}\ud \gamma(x,y)\right)^{1/p},
 \end{align*}
 where $\mu_{1}, \mu_{2}\in \calP_{p}(\calX)$, and $\Gamma(\mu_{1},\mu_{2})$ denotes the set of all couplings of $\mu_{1}$ and $\mu_{2}$, i.e., joint probability measures on $\calX\times\calX$ with marginals $\mu_{1}$ and $\mu_{2}$.
\end{Definition}

For a graph $\calG$, the signal sample space is $\calX=\bbR^N$ equipped with the metric distance $d(x,y)=\norm{x-y}_2$. The corresponding Wasserstein space $\calP_{p}(\bbR^N)$ consists of all probability measures on $\bbR^N$ with finite $p$-th moments, i.e., $\int_{\bbR^N} \norm{x}_2^p \ud\mu(x) < \infty$.

\begin{Definition}[\glspl{GDS}]
\label{def.gds}
A graph distribution-valued signal is a Borel probability measure $\mu \in \calP_{p}(\bbR^{N})$, where $\calP_{p}(\bbR^{N})$ denotes the $p$-Wasserstein space of probability measures on $\bbR^N$, the graph signal sample space. The space $\calP_{p}(\bbR^{N})$ is referred to as the space of \glspl{GDS}.
\end{Definition}

The Wasserstein distance measures the minimal ``work'' required to transport one probability distribution to another, where work is the product of the mass and distance transported. Equipped with $W_{p}$, the space $\calP_{p}(\bbR^{N})$ is a complete and separable \cite{Vil09}. While computing $W_p$ for general measures is challenging, closed-form solutions exist in special cases.

\begin{Example} \label{eg:imd}
\begin{enumerate}[(a)]
    \item \label{it:imd} If $\mu_1 = \delta_x$ and $\mu_2 = \delta_{y}$, 
    then $W_{2}(\delta_x,\delta_y) = d(x,y)$. Thus, the space of traditional graph signals $\bbR^N$ embeds isometrically into the space of \gls{GDS}s $\calP_{2}(\bbR^N)$. 
    \item If $\mu_1= \calN(\bm{m}_1,\bSigma_1)$ and $\mu_2= \calN(\bm{m}_2,\bSigma_2)$ are two non-degenerate Gaussian distributions on $\bbR^N$ with means $\bm{m}_1,\bm{m}_2$ and covariance matrices $\bSigma_1,\bSigma_2$ respectively. Then the Wasserstein distance is given by
    \begin{align*}
        &W_{2}(\mu_1,\mu_2)^2  = \norm{\bm{m}_1-\bm{m}_2}_{2}^2\\ 
        &\qquad + \Tr\left(\bSigma_1+\bSigma_2-2\left(\bSigma_2^{1/2}\bSigma_1\bSigma_2^{1/2}\right)^{1/2}\right).
    \end{align*}
    Therefore, minimizing the discrepancy between two distributions in the Wasserstein space involves matching not only their means but also their covariances.
\end{enumerate}
\end{Example}

\section{GDS framework and its application}
\label{sec:gds_framework}

In this section, we introduce a signal processing framework for \glspl{GDS}, generalizing traditional \gls{GSP}. We establish a dictionary linking classical GSP concepts to their GDS counterparts and then illustrate its use through a graph filter learning application.

\subsection{Signal processing for GDSs}

\subsubsection{GDS Fourier transform} 

Recall that in traditional \gls{GSP}, the \gls{GFT} is defined as an orthogonal change of basis with respect to the eigenbasis $\bU_{\calG}$ of a graph shift operator \gls{GSO} $\bS_{\calG}$. We extend this notion to \glspl{GDS} as follows.

\begin{Definition}[\gls{GDS} Fourier transform]
For a \gls{GDS} $\mu \in \calP_p(\bbR^N)$, the \emph{\gls{GDS-FT}} is the pushforward measure
\begin{align}
    \hat{\mu}:= \left(\bU\T_{\calG}\right)_{\#} \mu,
\end{align}
that is, for any Borel set $ B \in \calB(\bbR^N)$, $\widehat{\mu}(B) = \mu\parens*{\parens*{\bU\T_{\calG}}^{-1}(B)}$.
\end{Definition}

Intuitively, the \gls{GDS-FT} transports a distribution from the vertex domain to the frequency domain. If $\mu = \delta_{\bx}$ is a Dirac measure, then $\widehat{\mu} = \delta_{\bU\T_{\calG} \bx}$, which recovers the classical \gls{GFT}. The inverse \gls{GDS-FT} is given by $\mu = (\bU_\calG)_{\#}\widehat{\mu}$.  

\subsubsection{GDS convolutional filtering}

In the classical case, graph convolutional filtering applies a linear transformation that amplifies, attenuates, or removes certain frequency components of a signal. The \gls{GDS} formulation generalizes this idea by transforming an entire probability distribution through the graph operator, thereby capturing how both the signal and its statistics propagate across the graph.

\begin{Definition}[\gls{GDS} convolutional filtering]
\label{def.GDS_filters}
Let $\bF_{\calG}$ be a graph convolutional filter (i.e., a polynomial of $\bS_{\calG}$). Then \emph{\gls{GDS} convolutional filtering} is the mapping
\begin{align}
    \calF_{\calG}: \calP_p(\mathbb{R}^N) \to \calP_p(\mathbb{R}^N), \quad \mu \mapsto (\bF_\calG)_{\#}\mu.
\end{align}
\end{Definition}

If $\mu = \delta_{\bx}$ is a Dirac measure, then $\calF_\calG(\mu) = \delta_{\bF_\calG \bx}$, which recovers the classical graph convolution $\bF_\calG \bx$.

Intuitively, \cref{def.GDS_filters} indicates that if $\mu$ encodes the uncertainty or statistical properties of a graph signal, then $(\bF_\calG)_{\#}\mu$ represents how these properties are transformed by the filter $\bF_\calG$. In the vertex domain, the filter redistributes probability mass across nodes, while in the spectral domain, it modifies the distribution of frequency components. This generalizes classical filtering to operate directly on the entire distribution.

As an illustration, let $\mu$ be a Gaussian distribution with mean vector $\bm{m}$ and covariance matrix $\bSigma$. After applying the filter, $(\bF_\calG)_{\#}\mu$ is also Gaussian, with mean $\bF_\calG \bm{m}$ and covariance $\bF_\calG \bSigma \bF\T_{\calG}$. This demonstrates that the filter modifies not only the central tendency of the signal, represented by its mean, but also its variability and dependencies, captured by the covariance. In this way, the classical notion of graph filtering is extended from deterministic signals to full probabilistic representations.

\subsection{Graph Filter Learning using GDSs}

Suppose each node $v_i \in \calV$, $i=1,\dots,N$, is associated with a marginal distribution $\mu_i(\cdot;\theta_i)\in\calP_{p}(\bbR)$, parameterized by $\theta_i$ and estimated from local observations. Since synchronous measurements across all nodes may not be available, direct estimation of the joint distribution is infeasible. We therefore construct a joint distribution $\mu_{\theta,\kappa}\in \calP_{p}(\bbR^{N})$ by combining the marginals with a copula density $c_\kappa$, which captures inter-node dependencies:
\begin{align*}
    \mu_{\theta,\kappa}(x)=c_{\kappa}\left(F_{1}(x_{1};\theta_{1}),\dots,F_{N}(x_{N};\theta_{N})\right) \prod_{i=1}^{N} \mu_i(x_i;\theta_i)
\end{align*}
where $F_i(\cdot;\theta_i)$ denotes the cumulative distribution function (CDF) of $\mu_i(\cdot;\theta_i)$, $\theta= (\theta_i)_{i=1}^N$, and $\kappa$ parameterizes the copula. By Sklar's theorem \cite{Sklar1959,Durante2013}, the marginals of $\mu_{\theta,\kappa}$ are exactly $\mu_i(\cdot;\theta_i)$, $i=1,\dots,N$.

In the \gls{GDS} framework, graph filtering is the pushforward of a distribution under a filter $\bF_\calG$, yielding $\mu_{\theta,\kappa}':= (\bF_\calG)_{\#}\mu_{\theta,\kappa}$. Our objective is to identify the graph filter $\bF_{\calG}$ and the copula density $c_{\kappa}$ such that the filtered distribution $\mu_{\theta,\kappa}'$ matches a target distribution $\mu^{\star}$ in the Wasserstain space:
\begin{align}
\begin{aligned}
\label{eq.pro_formu}
    \min_{\bF_{\calG},\kappa} &\ W_{p}((\bF_\calG)_{\#}\mu_{\theta,\kappa},\mu^{\star})\\
    \ST &\ (\pi_i)_{\#} \mu_{\theta,\kappa} = \mu_i, \quad i=1,\dots,N,
\end{aligned}
\end{align}
where $\pi_i$ denotes the projection onto the $i$-th coordinate. This problem formulates graph filter learning for GDSs, where marginals represent local uncertainty, the copula models dependencies, and the learned filter aligns the overall distribution with the desired global structure.

\subsubsection{Copula-based GDS filter learning}

To make the optimization problem in \cref{eq.pro_formu} tractable, we adopt a Gaussian parameterization. Each node $v_i$ is assigned a marginal distribution $\mu_i=\calN(m_i,\sigma_i^2)$, where $m_i$ and $\sigma_i^2$ are estimated from local observations. Dependencies across nodes are modeled using a Gaussian copula with correlation matrix $\bm{R}$, yielding a joint distribution $\mu =\calN(\bm{m},\bm{\Sigma})$, where $\bm{m}=(m_1,\dots,m_N)^{\T}$ and $\bm{\Sigma}=\bm{DRD}$ with $\bm{D}=\mathrm{diag}(\sigma_1,\dots,\sigma_N)$. Applying a graph filter $\bF_\calG$ to $\mu$ yields another Gaussian distribution:
\begin{align*}
    (\bF_\calG)_{\#}\mu =\calN(\bF_{\calG}\bm{m}, \bF_{\calG}\bm{\Sigma}\bF\T_{\calG} ).
\end{align*}

Assuming the target distribution is also Gaussian, i.e., $\mu^{\star}=\calN(\bm{m}^{\star},\bm{\Sigma}^{\star})$, the squared Wasserstein distance between them has a closed-form expression. The learning problem in \cref{eq.pro_formu} reduces to:
\begin{align}
\begin{aligned}
\label{eq.opt}
\min_{\bF_{\calG}, \bm{R}} &\ \norm{\bF_{\calG}\bm{m}-\bm{m}^{\star}}_{2}^{2} + \Tr\Big(\bF_{\calG}\bm{\Sigma}\bF\T_{\calG} +\bm{\Sigma}^{\star}\\
    &\qquad\qquad- 2 \big( (\bm{\Sigma}^\star)^{1/2}\bF_{\calG}\bm{\Sigma}\bF_{\calG}^{\T}(\bm{\Sigma}^\star)^{1/2} \big)^{1/2}\Big)\\
\ST &\ \bm{R}=\bm{R}\T,\quad \bm{R}\succeq \mathbf{0}, \quad \diag(\bm{R})=\bone.
\end{aligned}
\end{align}
We solve this problem via alternating minimization over $\bF_{\calG}$ and $\bm{R}$ using gradient descent \cite{Lecun1998}. The full procedure is summarized in \cref{alg:gds_filter_learning}, referred to as the \emph{Copula-based GDS (GDS-Cop)} graph filtering learning algorithm.  

The optimized graph filter $\widetilde{\bF}_{\calG}$ obtained from solving \cref{eq.opt} can be directly applied for downstream prediction tasks. For example, if the target distribution is that of the signal in a future time period, given an input signal matrix $\bX \in \Real^{N \times T}$, the predicted future signal is computed as $\widehat{\bX} = \widetilde{\bF}_{\calG} \bX$. 

\begin{algorithm}[t]
\caption{Copula-based GDS graph filter learning algorithm}
\label{alg:gds_filter_learning}
\begin{algorithmic}[1]
\State Input: Target mean $\bm{m}^\star$ and covariance $\bm{\Sigma}^\star$, marginal means $\bm{m}$, variances $\bm{D} = \mathrm{diag}(\sigma_1, \dots, \sigma_N)$, learning rates $\eta_1$, $\eta_2$, convergence tolerance $\epsilon$, positive threshold $\delta$.
\State Output: Optimized $\widetilde{\bF}_{\calG}$
\State Initialize $\bF_{\calG}^{(0)}$, $\bm{R}^{(0)}$, $u = 0$
\Repeat
    \State $\bm{\Sigma}^{(u)} \gets \bm{D} \bm{R}^{(u)} \bm{D}$
    \State $\calL^{(u)} \gets \cref{eq.opt}$ 
    \State $\bF_{\calG}^{(u+1)} \gets \bF_{\calG}^{(u)} - \eta_1 \nabla_{\bF_{\calG}} \calL^{(u)}$
    \State $\bm{R}^{(u+1)} \gets \bm{R}^{(u)} - \eta_2 \nabla_{\bm{R}} \calL^{(u)}$
    \State Project $\bm{R}^{(u+1)}$ to a valid correlation matrix:
    \State $\bm{R}^{(u+1)} \gets \tfrac{1}{2} \left( \bm{R}^{(u+1)} + \bm{R}^{(u+1){\T}} \right)$ \hfill \Comment{Symmetrize}
    \State $\bm{R}^{(u+1)} \gets \bm{V} \diag(\max(\bm{\lambda}, \delta)) \bm{V}^{\T}$ \hfill \Comment{PSD projection}
    \State $\bm{S} \gets \diag\left( \sqrt{ \diag(\bm{R}^{(u+1)}) } \right)$
    \State $\bm{R}^{(u+1)} \gets \bm{S}^{-1} \bm{R}^{(u+1)} \bm{S}^{-1}$ \hfill \Comment{Normalize}
    \State $\diag(\bm{R}^{(u+1)}) \gets \mathbf{1}$ \Comment{Set diagonal}
    \State $u \gets u + 1$
\Until{$\norm{\calL^{(u)} - \calL^{(u-1)}} \le \epsilon$}
\end{algorithmic}
\end{algorithm}

\section{Numerical Experiments}

\subsection{Dataset and Experiment setup}

We evaluate \emph{GDS-Cop} on a COVID-19 dataset from The New York Times. \footnote{\url{https://github.com/nytimes/covid-19-data}} We use records for California’s $N=58$ counties from July 29, 2020, to August 1, 2022, splitting them into a training set $\calM_{\mathrm{train}}$ (July 29, 2020–July 30, 2021) and a test set $\calM_{\mathrm{test}}$ (July 31, 2021–August 1, 2022). Counties are nodes with edges defined by geographic adjacency. Both training and test sets are evenly partitioned into $\calS$ non-overlapping time windows, i.e., $\calM_{\mathrm{train}} = \{ f_{\mathrm{train}}(\calV, \calT_s)\}_{s=1}^{\calS}$, where $f_{\mathrm{train}}(\calV, \calT_s) $ denotes the number of cases reports across $\calV$ during the $s$-th time window $\calT_s$. Each window $\calT_{s}$ consists of $\calW$ consecutive days, so that the window size is given by $\calW=\abs{\calT_{s}}$.

During training, the target statistics, namely mean $\bm{m}^\star$ and covariance $\bm{\Sigma}^\star$, are estimated from the subsequent window $f_{\mathrm{train}}(\calV, \calT_{s+1})$ following the input window $f_{\mathrm{train}}(\calV, \calT_{s})$. The marginal mean $m_{i}$ and variance $\sigma_{i}$ for each node $v_i$ are computed from its local observations $f_{\mathrm{train}}(v_i, \calT_s)$. After optimizing $\widetilde{\bF}_{\calG}$ using \cref{eq.opt}, we evaluate it on the test set. For each test window $s = 1, \dots, \calS - 1$, the prediction is computed as
    $\widehat{f}_{\mathrm{test}}(\calV, \calT_{s+1}) = \widetilde{\bF}_{\calG} f_{\mathrm{test}}(\calV, \calT_s)$,
and the prediction error is evaluated using the relative squared error (RSE), defined by
\begin{align*}
   \text{RSE}^{(s)}=\frac{\norm{\widehat{f}_{\mathrm{test}}(\calV, \calT_{s+1})-f_{\mathrm{test}}(\calV, \calT_{s})}_{F}^{2}}{\norm{f_{\mathrm{test}}(\calV, \calT_{s})}_{F}^{2}}. 
\end{align*}
The final performance is measured by the average RSE across all prediction windows, defined as $\text{ARSE} = \frac{1}{\calS-1}\sum_{s=1}^{\calS-1} \text{RSE}^{(s)}$.

\subsection{Performance evaluation}
We compare \emph{GDS-Cop} againts the following graph filtering learning methods for graph signal prediction:
\begin{enumerate}[label=\alph*)]
    \item \textbf{GSP-LS} (Least-Squares Graph Filtering) \cite{Shu13, San13}: The graph filter $\bF_{\calG}$ is learned by minimizing the Frobenius norm of the difference between the filtered signals and the target signals, i.e.,
        $\min_{\bF_{\calG}} \norm{\bF_{\calG} \bX - \bX^\star}_F^2$.
    \item \textbf{GSP-RLS} (Regularized Least-Squares Graph Filtering) \cite{Ramirez2021,Isufi2024}: The graph filter $\bF_{\calG}$ is learned by minimizing the same Frobenius norm objective as in GSP-LS, but with an additional $\ell_1$-norm regularization term to promote sparsity in the filter coefficients, i.e., 
        $\min_{\bF_{\calG}} \norm{\bF_{\calG} \bX - \bX^\star}_F^2 + \lambda \norm{\bF_{\calG}}_1$,
    where $\lambda > 0$ controls the sparsity level.
    \item \textbf{GSP-LSCM} (Least-Squares and Covariance-Matching Graph Filtering): This method extends GSP-LS by naively introducing a covariance-matching term into the objective. The graph filter $\bF_{\calG}$ is learned by minimizing a combination of data-fitting and covariance alignment losses, i.e., 
    \begin{align*}
        \min_{\bF_{\calG}} \norm{\bF_{\calG} \bX - \bX^\star}_F^2 + \lambda \norm{\bF_{\calG} \bm{\Sigma}_{\bX} \bF_{\calG}^{\T} - \bm{\Sigma}_{\bX^\star}}_F^2,
    \end{align*}
    where $\bm{\Sigma}_{\bX}$ and $\bm{\Sigma}_{\bX^\star}$ are the empirical covariances of the input and target signals, respectively, and $\lambda >0$ is a trade-off parameter that balances the two terms.
    \item \textbf{GSP-LEV} (Log-Evidence Maximization with Heat Kernel Mixture) \cite{Kwak2021}: The graph filter is modeled as a convex combination of $K$ heat diffusion kernels, i.e., $\bH^{\calG}(\calT) = \sum_{\tau \in \calT} \pi^{(\tau)} \bH^{\calG}(\tau)$ with $\bH^{\calG}(\tau) = e^{-\tau \bL_{\calG}}$, where $\calT$ is a predefined set of diffusion scales and $\pi^{(\tau)} \geq 0$, $\sum_{\tau\in\calT} \pi^{(\tau)} = 1$. The parameters $\pi_{\tau}$, along with precision terms $\alpha$ and $\gamma$, are learned by maximizing the log-evidence: 
    \begin{align*}
        \max_{\pi, \alpha, \gamma} \log \mathcal{N}\left( \bX^{\star} \mid \bH^{\calG}(\calT)\bX,\, \alpha^{-1} \bI + \gamma^{-1} \bX \bX^{\T} \right).
    \end{align*}
\end{enumerate}
For a fair comparison, the graph filter $\bF_{\calG}$ is parameterized as a 2-order Chebyshev polynomial with three filter coefficients, i.e., $\bF_{\calG}=\sum_{k=0}^{2} \theta_{k} T(\bL_{\calG})$, and this parameterization is used consistently across all methods, including GSP-LS, GSP-RLS, GSP-LSCM, and the proposed GDS-Cop.

\begin{figure}[!h]
    \centering
    \includegraphics[width=1.025\columnwidth]{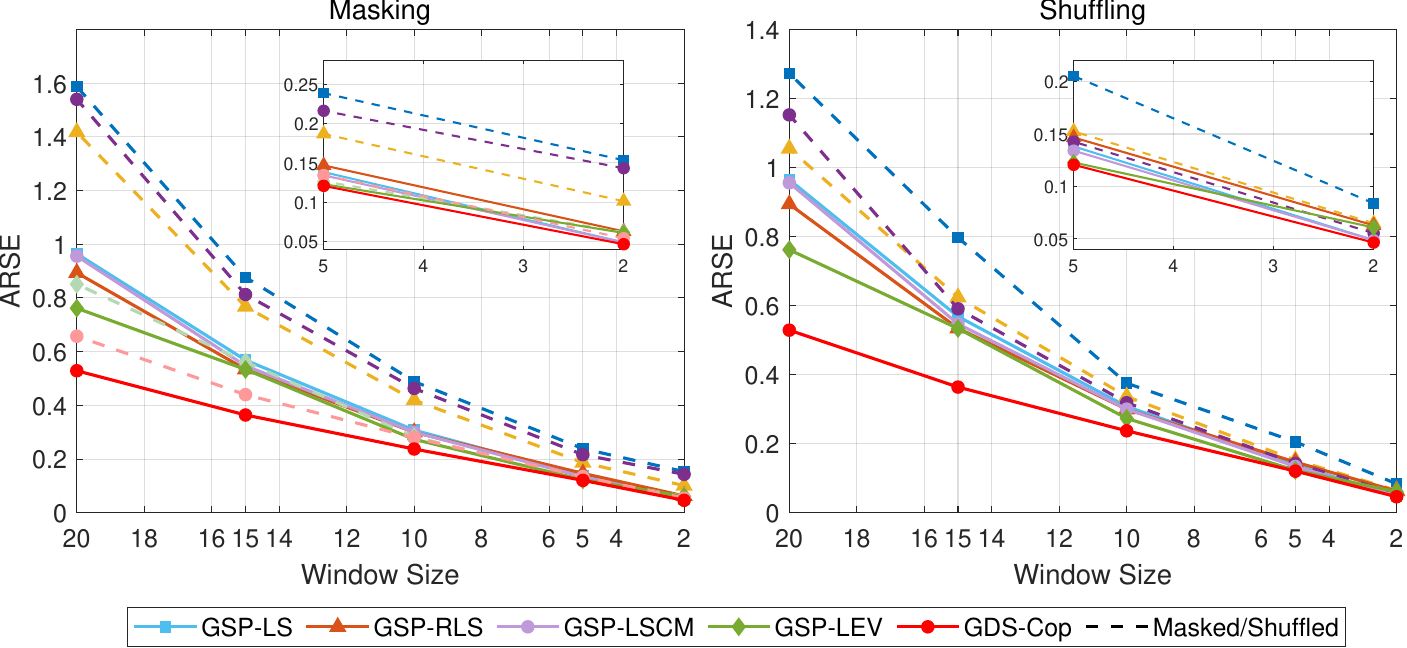}
        \begin{subfigure}{0.49\columnwidth}
        \phantomsubcaption
        \label{fig:masking}
        \centering \small (a) Non-masking VS. Masking 
    \end{subfigure}
    \hfill
    \begin{subfigure}{0.49\columnwidth}
        \phantomsubcaption
        \label{fig:shuffling}
        \centering \small (b) Non-shuffling VS. Shuffling
    \end{subfigure}
    \caption{ARSE with respect to different window sizes under masking and shuffling settings.}
    \label{fig:covid_RSE}
\end{figure}

To validate that the GDS framework relies on neither complete observations nor strict temporal correspondence, we consider two stress-test settings: \textbf{masking} and \textbf{shuffling}. In masking, a Bernoulli mask is applied so that, for each graph signal (column), entries are independently observed with probability drawn from $[0.6,0.9]$. Each experiment is repeated 10 times with different masks, and results are averaged. From \cref{fig:masking}, we observe that GSP-LS, GSP-RLS, and GSP-LSCM degrade markedly due to their reliance on the complete observation assumption, while GSP-LEV and \emph{GDS-Cop} remain relatively robust since their distribution-matching formulations do not require full observations. Under shuffling, training data within each window is randomly permuted. Results in \cref{fig:shuffling} are averaged over 10 runs. GSP-LS, GSP-RLS, and GSP-LSCM suffer worse accuracy because they depend on strict temporal alignment, whereas GSP-LEV and \emph{GDS-Cop} remain stable. This resilience underscores the suitability of \emph{GDS-Cop} for real-world scenarios with irregular or misaligned data. From \cref{fig:covid_RSE}, we also observe that \emph{GDS-Cop} consistently achieves the lowest ARSE, with its advantage most evident for larger windows. As $\calW$ decreases, all methods improve due to the shorter horizon, yet \emph{GDS-Cop} remains the most accurate and stable. 


\section{Conclusion}
We propose a \gls{GDS} framework that generalizes classical graph signal processing by modeling signals as probability distributions in the Wasserstein space. This approach enables the principled handling of uncertainty and stochasticity, while strictly encompassing traditional GSP as a special case. We establish a systematic correspondence between core GSP concepts and their GDS analogues, including Fourier transforms and filtering. Experimental results on graph filter learning tasks demonstrate the predictive benefits and robustness of the proposed framework, highlighting its potential for advancing GSP theory and supporting applications involving uncertain or irregular graph-structured data.

\end{document}